\documentclass{article}

\usepackage[numbers]{natbib}
\usepackage[final]{neurips_distshift_2021}

\usepackage[utf8]{inputenc} 
\usepackage[T1]{fontenc}    

\usepackage{url}            
\usepackage{booktabs}       
\usepackage{amsfonts}       
\usepackage{nicefrac}       
\usepackage{microtype}      

\usepackage{graphics} 
\usepackage{epsfig} 
 \usepackage{array}
\usepackage[column=0]{cellspace}
\usepackage{hhline}
\usepackage{multirow, makecell}
\usepackage{times} 
\usepackage{amsmath} 
\usepackage{amssymb}  
\usepackage{cancel}
\usepackage{wasysym}
\usepackage{mathtools,tikz}
\usepackage{xhfill}
\usepackage[ruled,vlined, linesnumbered]{algorithm2e}
\usepackage{wrapfig}
\newcommand{\ind}{\perp\!\!\!\!\perp} 
\providecommand\rightarrowRHD{\relbar\joinrel\mathrel\RHD}
\providecommand\dashbiarrow{\LHD\makebox[0.7em]{\xdotfill{.4pt}}\RHD}
\providecommand\hollowdasharrow{\makebox[0.7em]{\xdotfill{.4pt}}\rhd}
\providecommand\bluearrow{\color{blue}\rightarrowRHD}
\providecommand\greenarrow{\color{green!50!black}\rightarrowRHD}
\providecommand\redarrow{\color{red!75!black}\rightarrowRHD}
\usepackage{hyperref}       
\usepackage{xcolor}         

\title{Causal-based Time Series Domain Generalization for Vehicle Intention Prediction}

\author{%
    Yeping Hu \\
    Department of Mechanical Engineering\\
    University of California, Berkeley\\
    \texttt{yeping\_hu@berkeley.edu} \\
    \And
    Xiaogang Jia \\
    Department of Computer Science\\
    University of Bristol\\
    \texttt{xiaogangjia\_hit@outlook.com} \\
    \AND
    Masayoshi Tomizuka \\
    Department of Mechanical Engineering\\
    University of California, Berkeley\\
    \texttt{tomizuka@berkeley.edu} \\
    \And
    Wei Zhan \\
    Department of Mechanical Engineering\\
    University of California, Berkeley\\
    \texttt{wzhan@berkeley.edu} \\

}

\begin{document}

\maketitle

\begin{abstract}
    Accurately predicting possible behaviors of traffic participants is an essential capability for autonomous vehicles. Since autonomous vehicles need to navigate in dynamically changing environments, they are expected to make accurate predictions regardless of where they are and what driving circumstances they encountered. Therefore, generalization capability to unseen domains is crucial for prediction models when autonomous vehicles are deployed in the real world. In this paper, we aim to address the domain generalization problem for vehicle intention prediction tasks and a causal-based time series domain generalization (CTSDG) model is proposed. We construct a structural causal model for vehicle intention prediction tasks to learn an invariant representation of input driving data for domain generalization. We further integrate a recurrent latent variable model into our structural causal model to better capture temporal latent dependencies from time-series input data. The effectiveness of our approach is evaluated via real-world driving data. We demonstrate that our proposed method has consistent improvement on prediction accuracy compared to other state-of-the-art domain generalization and behavior prediction methods.
\end{abstract}

\section{Introduction}

Researchers have developed various machine learning algorithms to enhance the accuracy of intention and motion prediction tasks for autonomous vehicles \cite{TNT, DESIRE, hu2018probabilistic, hu2020scenario, hu2019multi, hu2019generic, Intentnet,  Precog, hu2018framework, Traffic}. Although these machine learning algorithms provide excellent prediction performance, a key assumption underlying the remarkable success is that the training and test data usually follow similar statistics. Otherwise, when test domains are unseen or Out-of-Distribution (OOD) \cite{OOD_AV}, the resulting train-test domain shift leads to significant degradation in prediction performance. 
Domain generalization (DG) aims to build models that are designed for increased robustness to domain-shift without requiring access to target domain data. Many researchers have been working on developing DG algorithms \cite{ghifary2015domain, li2018domain, li2018learning, balaji2018metareg, carluccidomain, cdann, causal_1, MatchDG, causal_2, mixup}, but most of these models are designed for tasks such as object recognition, semantic segmentation, image classification, and disease detection. In this paper, our goal is to design a DG algorithm for vehicle intention prediction task which is different from typical DG tasks in three aspects: (1) the input is numerical time-series data instead of static image; (2) each input data point contains information of multiple agents (e.g. vehicles) with interactions involved, instead of a single agent (e.g. patient) with no interaction; (3) factors that cause domain shift are ambiguous for our task, in contrast to clear domain shift variables for other DG tasks (e.g. image rotation angle or style for image classification task). Therefore, directly applying state-of-the-art DG approaches may not be suitable for our problem setting and could result in undesirable performance. It is thus necessary to develop a domain generalization model that considers the discrepancies in the  aforementioned  aspects  and  is  applicable  for  vehicle intention prediction task.
In this paper, a causal-based time series domain generalization (CTSDG) model for vehicle intention prediction is proposed. To the best of our knowledge, this is the first work trying to tackle time series domain generalization problem for vehicle intention prediction task. The effectiveness of our approach is evaluated on real-world driving data under different scenarios. Our method also shows consistent improvement on prediction accuracy compared to other state-of-the-art methods.

\section{Method}

\subsection{Notations}
Let $\mathcal{X}$ be the input space and $\mathcal{Y}$ the label space, a \textit{domain} is defined as a joint distribution $P_{XY}$ on $\mathcal{X} \times \mathcal{Y}$. In the context of domain generalization, we assume to have access to $K$ domains, $\mathcal{D} = \{D_k\}^{K}_{k=1}$, each associated with a joint distribution $P_{XY}^{k}$. In each domain, data-label pairs are sampled from a dataset $D_k = \{(\mathbf{x}_i^{(k)},\mathbf{y}_i^{(k)})\}_{i=1}^{N_k}$ with $(\mathbf{x}_i^{(k)},\mathbf{y}_i^{(k)}) \sim P_{XY}^{k}$, where $N_k$ is the number of labeled data points in the $k$-th domain and $\mathbf{x}_i = (x_t^i)_{t=1}^{T} = ([\xi_t^{A,i}, \xi_t^{B,i}])_{t=1}^T$ denotes a multivariate time series with $x_t^i \in \mathbb{R}^{H \times 2}$. For a given vehicle, we use $\xi$ to represent historical trajectories. In this work, given historical trajectories of interacting vehicle pairs (i.e. car A and car B), the domain generalization task is to learn an intention predictor $f$ (i.e. predict interaction outcomes such as pass/yield) that generalizes well to unseen target domains $k' \notin D_k$ where $P_{XY}^{(k')} \neq P_{XY}^{(k)}$, $\forall k \in \{1,\dots,K\}$.


\subsection{Learning Invariant Representation from Causal Model}
\begin{wrapfigure}{r}{0.5\textwidth}
  \vspace{-15pt}  
  \begin{center}
    \includegraphics[width=0.48\textwidth]{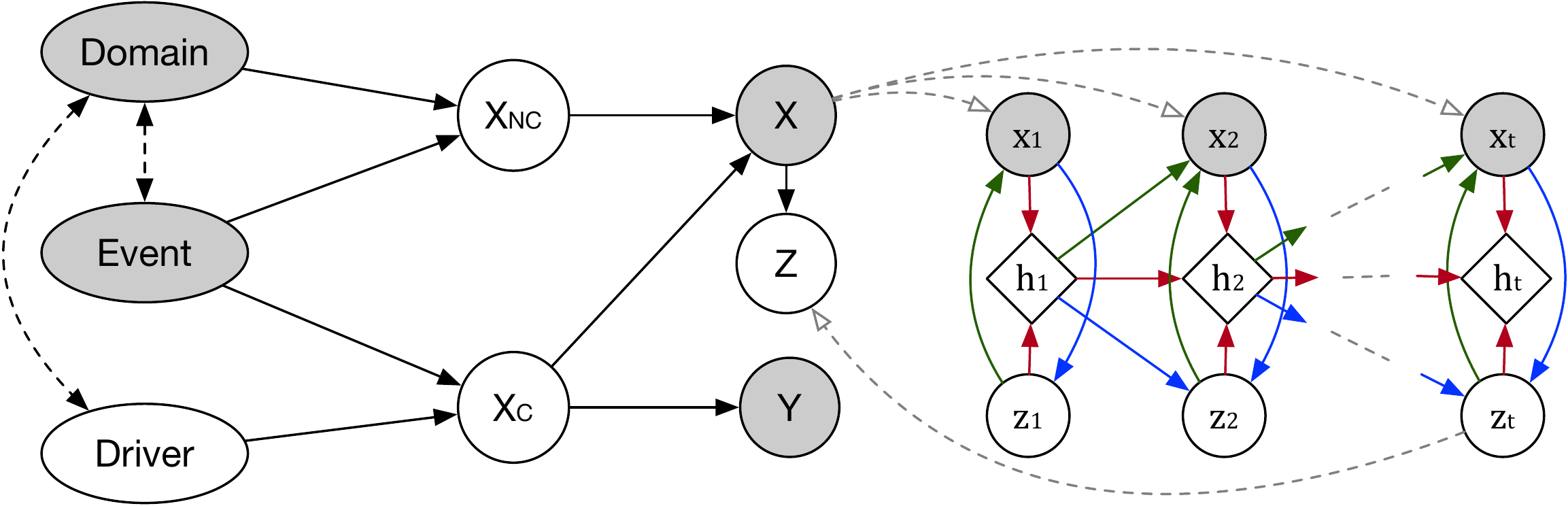}
  \end{center}
  \caption{Illustration of the overall CTSDG framework. Observed quantities are shown as shaded nodes; nodes of latent quantities are transparent. A directed black edge ($ \protect\rightarrowRHD$) denotes a causal relationship; dashed bidirectional edges ($ \protect\dashbiarrow$) denote correlation; hollow dashed arrows ($ \protect\hollowdasharrow$) denote an inclusion relationship. Blue lines ($ \protect\bluearrow$) denote the inference process, $q(z_t|x_{ \leq t}, z_{< t})$; green lines ($ \protect\greenarrow$) represent the generation process, $p(x_t|z_{ \leq t}, x_{<t})$; red lines ($ \protect\redarrow$) show the recurrence process where $h_t$ is informed by $h_{t-1}$, which is informed by $z_{t-1}$ and $x_{t-1}$.}
  \vspace{-10pt}
  \label{fig:framework}
\end{wrapfigure}
In this work, we propose a causal view of data generating process under vehicle interaction settings. More specifically, we utilize a structural causal model (SCM) based on domain knowledge to describe how interactive trajectories are generated by any vehicle pair (see Figure \ref{fig:framework}). The constructed causal structure provides us with causal relations between input features and the outcome, which is used to learn the intention predictor. The detailed meaning of each node variable is as follow: \textbf{Domain ($D$)} contains different combinations of map properties such as road topology, speed limit, and traffic rules; \textbf{Event ($E$)} denotes an observable variable relate to the event of two-vehicle interaction, which includes information such as initial interaction states and the length of interaction; \textbf{Driver ($O$)} denotes an unobservable variable relate to each driver's driving preferences or style, such as the aggressiveness level and whether he/she obeys traffic rules; \textbf{Causal features} ($X_C$) represent high-level time-related causal features derived from drivers' personal behavior $O$ and their interaction status $E$, which are used by humans to label the intention; \textbf{Non-causal features} ($X_{NC}$) represent domain dependent features which contain not only domain-specific information $D$, but also event $E$. For example, interaction event information such as when two vehicles start noticing each other or when the road negotiation begins, is related to the road geometry and traffic rule; \textbf{Input Data} ($X$) is vehicle interactive trajectories that contain sequential multivariate data, and is constructed from a mixture of causal $X_C$ and non-causal $X_{NC}$ features; \textbf{Latent variable} ($Z$) denotes time-related latent representations extracted from time-series vehicle interaction data; \textbf{Label} ($Y$) is the vehicle intention label. 

\textbf{Invariance Condition.} According to Figure~\ref{fig:framework}, $X_C$ is the node that causes $Y$, and by d-separation \cite{pearl2009causality}, the intention label is independent of domain conditioned on $X_C$, $Y \ind D | X_C$\footnote{The notation $Y \ind D | X_C$ stands for the conditional independence relationship $P(Y = y, D = k| X_C = \mathbf{x}_c) = P(Y = y| X_C = \mathbf{x}_c)P(D = k|X_C = \mathbf{x}_c)$.}. In other word, if such a $X_C$ can be found, then the
distribution of Y conditional on $X_C$ is invariant under transferring from the source domains to the target domains. Therefore, our intention prediction task is to learn $y$ as $h(\mathbf{x}_c)$ where $h: \mathcal{C} \rightarrow \mathcal{Y}$. However, since $X_C$ is unobserved, we need to learn $y$ through observed trajectories $X$. Specifically, we utilize a representation function $q: \mathcal{X} \rightarrow \mathcal{Z}$ to map the input space to a latent space, and a hypothesis function $\phi: \mathcal{Z} \rightarrow \mathcal{C}$ to map the latent space to $X_C$. Together, $h(\phi(q(\mathbf{x})))$ leads to the desired intention predictor $f: \mathcal{X} \rightarrow \mathcal{Y}$ and the corresponding prediction loss can be written as:
\begin{equation}\label{eq:L_y}
    \mathcal{L}_y = \mathcal{L}_\alpha(h(\phi(q(X))), Y),
\end{equation}
where $\mathcal{L}_\alpha$ is the classification loss such as a binary or categorical cross-entropy. In addition, by d-separation, $X_C$ also needs to satisfy an invariance condition: $X_C \ind D|\{E, O\}$, which means $X_C$ does not change with different domains when both event and driver information remain the same. However, driver information is unobservable and in many dataset there may not be an exact match based on a same interaction event across domains. Alternatively, we assume that the distance over $X_C$ between same-class inputs from different domains is bounded, which provides an alternative invariance condition that is consistent with the conditional independencies of $X_C$. 
Therefore, we would like to minimize the following objective along with the prediction loss:
\begin{equation}\label{eq:L_r}
    \mathcal{L}_r = \sum_{\Omega(\mathbf{x}_j,\mathbf{x}_m)=1; k_j \neq k_m}\mathcal{L}_\beta(\phi(q(\mathbf{x}_j^{(k_j)})), \phi(q(\mathbf{x}_m^{(k_m)}))),
\end{equation}
where $\mathcal{L}_\beta$ is the distance metric such as $\ell_2$, and $\Omega: \mathcal{X} \times \mathcal{X} \rightarrow \{0, 1\}$ is a match function such that pairs having $\Omega(\mathbf{x}_j, \mathbf{x}_m) = 1$ have low difference in their causal features.

\textbf{Contrastive Representation Learning.} To optimize $\mathcal{L}_r$, we first need to learn a proper match function $\Omega$ used in Eq.~(\ref{eq:L_r}). Specifically, we optimize a contrastive representation learning loss that minimizes distance between same-class inputs from different domains. We regard positive matches as two inputs from the same class but different domains, and negative matches as pairs with different classes. Then the loss function for every positive match pair $(j, m)$ in a sampled mini-batch $\mathcal{B}$ is defined as:
\begin{eqnarray}\label{eq:contrast_loss}
    \ell^{j,m}_{con} = -\log\frac{\mathrm{exp}(s_{cos}(\mathbf{x}_j, \mathbf{x}_m)/\tau)}{\mathrm{exp}(s_{cos}(\mathbf{x}_j, \mathbf{x}_m)/\tau) + \sum_{i=1, y_i \neq y_j}^{|\mathcal{B}|}\mathrm{exp}(s_{cos}(\mathbf{x}_j, \mathbf{x}_i)/\tau)},
\end{eqnarray}
where $s_{cos}(\mathbf{x}_a, \mathbf{x}_b) = \phi(q(\mathbf{x}_a))^T\phi(q(\mathbf{x}_b))/\|\phi(q(\mathbf{x}_a))\| \|\phi(q(\mathbf{x}_b))\|$ is the inner product of two $\ell_2$-normalized vectors, $\tau$ is a temperature scaling parameter, and $|\mathcal{B}|$ is the batch size. Inspired from \cite{MatchDG}, for this unsupervised contrastive learning process, we initialize $\Omega$ with a random match based on classes and keep updating $\Omega$ by minimizing the contrastive loss (\ref{eq:contrast_loss}) until convergence.

\subsection{Capturing Temporal Latent Dependencies}

From our proposed causal model in Fig.\ref{fig:framework}, we also observe that $Z \ind D | X$, meaning the latent representation is independent of domain $D$ conditioned on input data $X$. Therefore, by learning the representation function $q$, we are able to extract a domain-invariant latent variable that represents the input space. Since the input is time-series data, the learned latent variable is expect to capture temporal latent information in the data. To explicitly model the dependencies between latent random variable across time steps, we integrate Variational Recurrent Neural Networks (VRNN) \cite{VRNN} into our CTSDG model. 
The VRNN contains a VAE at every time step and these VAEs are conditioned on previous auto-encoders via the hidden  state variable $h_{t-1}$ of an RNN. 
In general, the objective function become a timestep-wise variational lower bound:
\begin{equation}\label{eq:L_v}
    \begin{split}
        \mathcal{L}_{v} = \mathbb{E}_{q(z_{\leq T}^i|x_{\leq T}^i)}[\sum_{t=1}^{T}(-\text{KL}(q(z_t^i|x_{\leq t}^i, z_{\leq t}^i)||p(z_t^i|x_{< t}^i, z_{< t}^i)) +\log p(x_t^i|z_{\leq t}^i, x_{\leq t}^i))],
    \end{split}
\end{equation}
where $q(z_t^i|x_{\leq t}^i, z_{\leq t}^i)$ is the inference model, $p(z_t^i|x_{< t}^i, z_{< t}^i))$ is the prior, $p(x_t^i|z_{\leq t}^i, x_{\leq t}^i)$ is the generative model, and $\text{KL}(\cdot||\cdot)$ refers to Kullback-Leibler divergence. 
\begin{table}[h!]
  \begin{center}
    \caption{Domain generalization results with prediction accuracy (\%).}
    \label{tab:FT_results}
    \resizebox{\textwidth}{!}{
    \begin{tabular}{c|ccccccc|ccc|c}
      \toprule 
      \multicolumn{1}{c|}{} &
	  \multicolumn{7}{c|}{\textbf{Traditional Baseline Methods}} &
	  \multicolumn{3}{c|}{\textbf{Behavior Prediction Methods}} &
	  \multicolumn{1}{c}{\textbf{Ours}} \\
      \midrule
      Source / Target & ERM & IRM & CCSA & Mixup & DANN & C-DANN & VRADA & {CVAE} & {GAN} & {GNN} & CTSDG \\
       &  \cite{erm} & \cite{irm} & \cite{ccsa} & \cite{mixup} & \cite{dann} & \cite{cdann} & \cite{vrada} & \cite{DESIRE} & \cite{social_GAN} & {\cite{NRI}} & \\
      \midrule 
      \midrule
      FT-1,2 / FT-3 & 75.97 (2.01) & 82.68 (3.91) & 78.03 (2.26) & 60.70 (16.61) & 77.47 (4.69) & 81.00 (4.87) & 76.35 (10.34) & {75.23 (3.27)} & {74.46 (2.36)} &  {76.34 (1.67)} & 86.03 (1.48) \\
      FT-1,3 / FT-2 & 97.28 (0.82) & 93.28 (3.83) & 96.68 (1.42) & 94.05 (3.15) & 95.66 (0.77)  & 95.06 (3.48) & 96.43 (0.51) & {95.32 (0.38)} & {94.98 (0.96)} &  {97.01 (0.53)} & 96.85 (0.15) \\
      FT-2,3 / FT-1 & 98.38 (0.72) & 98.76 (0.20) & 98.29 (0.62) & 91.56 (7.81) & 99.23 (0.34)  & 99.19 (0.45) & 98.12 (0.97) & {96.85 (0.63)} & {99.10 (0.13)} & {98.77 (0.80)}  & 98.85 (0.13) \\
      \midrule
      \multicolumn{1}{c|}{Average} & 90.54 & 91.58 & 91.00 & 82.10 & 90.79 & 91.75 & 90.30 & {89.13} & {89.51} & {90.71} & \textbf{93.91}\\
      \midrule 
      \midrule
      FT-1,2 / ZS & 72.71 (3.02) & 75.60 (3.99) & 74.64 (2.61) & 69.08 (6.25) & 77.05 (4.83)  & 76.40 (5.55) & 83.81 (3.57) & {65.32 (2.34)} & {78.02 (5.34)} &  {76.31 (4.49)} & 84.06 (3.54) \\
      FT-1,3 / ZS & 80.67 (2.74) & 80.43 (1.92) & 80.92 (5.63) & 79.23 (1.50) & 82.37 (2.54)  & 84.06 (4.35) & 77.54 (6.91) & {70.77 (4.24)} & {78.26 (4.75)} &  {77.23 (4.67)} & 84.78 (3.49) \\
      FT-2,3 / ZS & 57.97 (0.73) & 62.80 (1.67) & 61.35 (2.74) & 60.62 (11.55) & 71.50 (13.87)  & 69.56 (10.67) & 68.11 (9.25) & {74.66 (6.03)} & {75.43 (1.51)} &  {75.01 (2.91)} & 77.54 (3.53)\\
      FT-1,2,3 / ZS & 80.43 (3.32) & 76.33 (3.72) & 78.02 (3.64) & 70.53 (15.92) & 82.61 (3.16)  & 82.37  (2.54) & 85.02 (2.93) & {73.43 (4.93)} & {83.57 (2.93)} &  {84.10 (3.22)} & 85.51 (2.66) \\
      \midrule
      Average & 72.95 & 73.79 & 73.73 & 69.87 & 78.38 & 78.10 & 78.62 & {71.04} &{78.82} & {78.16} & \textbf{82.97}\\
      \bottomrule 
    \end{tabular}}
  \end{center}
\end{table}
\subsection{Overall Algorithm}
Combining the joint optimization problem of equation (\ref{eq:L_y}), (\ref{eq:L_r}), and (\ref{eq:L_v}) leads to our CTSDG model. The complete objective function we aim to optimize is:
\begin{equation}\label{eq:L}
    \mathcal{L} = \mathcal{L}_y + \gamma \mathcal{L}_v + \lambda \mathcal{L}_r,
\end{equation}
where $\gamma$ and $\lambda$ are hyperparameters that control training balance among three losses for better performance. In order to prevent the classification loss from interfering learning domain-invariant representations, we learn the match function $\Omega$ before optimizing the overall objective function (\ref{eq:L}). Also, since the causal feature $X_C$ is a function of latent variable $Z$ which is extracted by the recurrent latent variable model, the objective function (\ref{eq:L_v}) needs to be jointly optimized while learning the matching function using Eq.(\ref{eq:contrast_loss}). 



\IncMargin{1.5em}
\SetKwInOut{Input}{Input}
\SetKwInOut{Output}{Output}

\begin{wrapfigure}{r}{0.55\textwidth}
  \vspace{-15pt} 
\begin{algorithm}[H]
\SetAlgoLined
\Indm
\Input{Source training domain $\mathcal{D} = \{D_k\}_{k=1}^{K}$; hyperparameters $\tau, \gamma, \lambda > 0$}  
\Output{Intention Predictor $f: \mathcal{X} \rightarrow \mathcal{Y}$ }
\Indp
Randomly split source domains $\mathcal{D}$ into disjoint train $\mathcal{D}_{tr}$ and validation $\mathcal{D}_{val}$;\\
Initialize $\Omega$ as a random match function;\\
 \While{not converged}{
  Sample a mini-batch from $\mathcal{D}_{tr}$;\\
  \eIf{epoch $<$ threshold}{
   Minimize (\ref{eq:contrast_loss}) + $\gamma$(\ref{eq:L_v});\\
   Update $\Omega$;
   }{
   Minimize (\ref{eq:L});
  }
 }
 \caption{Causal-based time-series domain generalization (CTSDG)}
\end{algorithm}
  \vspace{-14pt}
  
\end{wrapfigure}

The pseudocode for our proposed CTSDG method is shown in Algorithm 1. We first initialize $\Omega$ as a random match function, where given each data point, we randomly select another data point with the same class from another domain. During each training epoch, we constantly update $\Omega$ based on the currently learned hypothesis function $\phi$. Specifically, given any data point $x$, we find the nearest neighbour of $x$ among other data points with the same class from another domain by calculating their distance using $\mathcal{L}_\beta$ defined in Eq.(\ref{eq:L_r}). After certain number of training epoch is reached, we further optimize the overall objective function (\ref{eq:L}) until convergence. Note that when calculating $\mathcal{L}_r$ in Eq.(\ref{eq:L_r}), we apply the updated match function $\Omega$ to select match pairs. 
\section{Experiments}



\begin{figure}[htbp]
	\centering
	\includegraphics[scale=0.4]{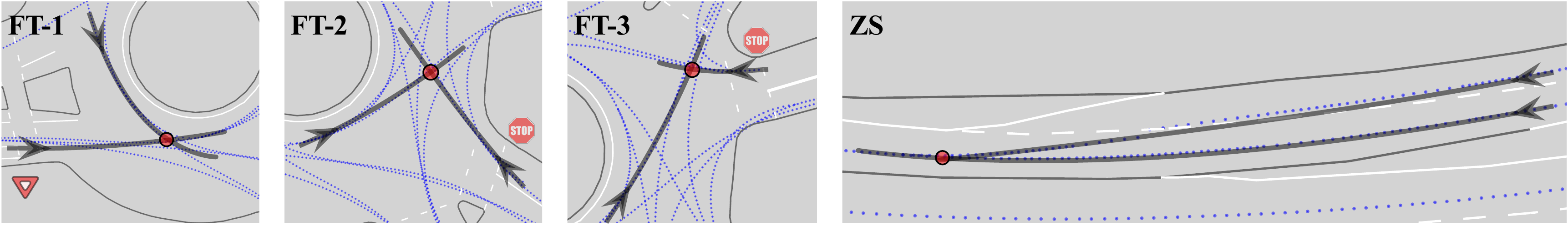}
	\caption{Illustration of selected domains for driving scenarios. Three domains selected from USA\_Roundabout\_FT are denoted as FT-1, FT-2, and FT-3. The one domain selected from CHN\_Merging\_ZS is denoted as ZS. Each black arrow line represents a reference path and red circles represent intersect points of reference paths.}
	\label{fig:nips_map}
\end{figure}

The experiment is conducted on the INTERACTION dataset \cite{dataset}, where two different scenarios are utilized: USA\_Roundabout\_FT and CHN\_Merging\_ZS. 
In this work, a domain is defined as a driving region that contains a unique combination of road topology, traffic rules, and speed limit information. Three domains are extracted from the FT scenario and one domain is extracted from the ZS scenario (see Figure \ref{fig:nips_map}). 
We compare our method with ten state-of-the-art methods, including seven traditional baseline methods for domain generalization tasks and three methods for behavior prediction tasks. 
To evaluate the domain generalization performance, we follow the typically used \textit{leave-one-domain-out} protocol \cite{li2017deeper}. Details related to data selection, implementation, and more in-depth evaluations on learned latent  representations  and  temporal  latent  dependencies can be found in supplementary materials. 

\textbf{Evaluation on USA\_Roundabout\_FT.} We first train all models on different combinations of source domains extracted in FT and report the test results on the remaining target domain in FT. From Table \ref{tab:FT_results}, we observe that our method achieves an average prediction accuracy of 93.91\% under all three train-test domain pairs and increases the accuracy by 2.54\% compared to the best-performed baseline model. 
ERM and GNN method slightly outperform CTSDG when FT-2 is the target domain, but their accuracy is 10.24\% and 11.26\% less than our method when tested on FT-3, respectively. 
As seen, our method has consistent prediction accuracy regardless of testing domains. The standard deviations of CTSDG are also lower than those of all other baseline methods, which further shows our model's efficacy as it converges to more stable local optima. 


\textbf{Evaluation on CHN\_Merging\_ZS.}
When we compare the domain in ZS with domains in FT, we notice larger difference in road topology as well as in speed limit. In other word, the joint distribution shift is more severe between domains from two different scenarios (i.e. ZS and FT) than domains within scenario FT, which makes domain generalization more challenging. Therefore, to evaluate the generalizability of each model under large domain shift, we select different domain combinations in FT for training and the single domain in ZS for testing. From Table \ref{tab:FT_results}, we observe that CTSDG consistently outperforms all baseline models across all source-target pairs with accuracy 1\% $\sim$ 6\% higher. The average prediction accuracy of our method increases by 5\%  compared to the best-performed baseline model. Also, when the model is trained using all three domains in FT, where the source domains have more diversity, learning based methods that do not specifically consider domain generalization approaches (i.e. ERM, CVAE, GAN, and GNN) have reasonable performance. However, the performance of these methods degrade drastically when training data are limited and lack of diversity. 
According to the results, we show the diminishing performance of traditional machine learning model to out-of-training distribution regimes and our proposed method is able to mitigate such problem while consistently outperforms other state-of-the-art methods.

\section{Extension to Trajectory Prediction Tasks}
In this paper, although we illustrate the performance of our proposed domain generalization method for vehicle intention prediction tasks, our CTSDG approach can also be directly or indirectly utilized for trajectory prediction tasks. The most straight forward way is to frame the trajectory prediction problem as classification over a diverse set of trajectories \cite{phan2020covernet}, which enables us to a) ensure a desired level of coverage of the state space, b) eliminate dynamically infeasible trajectories, and c) avoid the issue of mode collapse. Moreover, the accuracy of intention classification plays an important role in regression tasks. As stated and shown in \cite{TNT, deo2018multi, goal_plan_predestrain, ITSC_ttlc},  goal/intent-conditioned trajectory forecasting can improve joint-agent and per-agent predictions, compared to unconditional forecast. Conditional forecasting can be also used for planning tasks, which effectiveness is demonstrated in \cite{Precog}. Therefore, having an accurate and domain generalizable intention/goal predictor is the prerequisite of developing state-of-the-art trajectory predictor and motion planner.
\section{Conclusions}
In this paper, we present a causal-based time-series domain generalization model (CTSDG) to address domain generalization problem for vehicle intention prediction tasks. We show that CTSDG has consistent improvements in terms of prediction accuracy compared to other state-of-the-art domain generalization and behavior prediction methods using real-world driving data. In the future, we expect to extend our work to address trajectory prediction problems and consider generalization of varying number of interacting vehicles. 

\bibliographystyle{unsrt}
\bibliography{neurips_distshift_2021}

\clearpage
\begin{center}
\textbf{\large Supplemental Materials}
\end{center}
\renewcommand{\thesubsection}{\Alph{subsection}}

\subsection{Related Works}
\label{related_works}

\textbf{Domain generalization}
aims to generalize models to unseen domains without knowledge about the target distribution during training. Different methods have been proposed for learning generalizable and transferable representations. A commonly-used strategy is to extract task-specific but domain-invariant features \cite{ghifary2015domain, li2018domain, cdann, causal_1, MatchDG, causal_2}. Ghifary et al. \cite{ghifary2015domain} learn multi-task auto-encoders to extract invariant features which are robust to domain variations. Li et al. \cite{li2018domain}
extend adversarial autoencoders by imposing maximum mean discrepancy measure to align multi-domain distributions. Li et al. \cite{cdann} consider the conditional distribution of label space over input space, and minimize discrepancy of a joint distribution. Recently, several works \cite{causal_1, MatchDG, causal_2} utilize causal structures to learn invariance factors, which have shown better model generalizability on unseen data, especially on data from different distributions than the train distribution. 

Dataset augmentation is another strategy that focuses on creating more out-of-domain samples by adversarially perturbing samples \cite{volpi2018generalizing}, leveraging self-supervised signals \cite{carluccidomain}, or utilizing convex combinations of input and output data \cite{mixup}. These data augmentation methods have shown effective for model generalization, but invalid samples could be generated and degrade the model performance if input data is vehicle trajectories that need to satisfy feasibility constraints. Meta-learning can also be applied in domain generalization, by creating meta-train and meta-test sets, and training a model using the meta-train set in such a way to improve the performance on the meta-test set \cite{finn2017model, li2018learning, balaji2018metareg}. While our proposed method can achieve promising results using standard training schemes, combining meta-learning with our approach could be an interesting future direction.

\textbf{Intention prediction in autonomous driving}
aims to predict high-level behaviors of road entities, which results are often used in downstream trajectory prediction modules. For example, works such as \cite{deo2018multi}, \cite{goal_plan_predestrain} and \cite{ITSC_ttlc} demonstrate that by predicting  goal  states  and  assuming  that  agents  navigate toward  those  goals  by  following  some  optimal  or  learned trajectories, the accuracy of prediction can be improved. Therefore, accurate intention prediction algorithm is essential for autonomous vehicles to smoothly navigate in dynamic environments. Zhang et al. \cite{zhang2013understanding} propose to model high level semantics in the form of traffic patterns through a generative model. Dong et al. \cite{pass_yield_merge_PGM} use a probabilistic graphical model to estimate the pass or yield probability under merging scenarios. Schulz et al. \cite{ITSC_DBN_intent} utilize the dynamic bayesian network method for routing prediction at intersections. Zyner et al. \cite{roundabout_RNN_intent_traj} obtain a probability distribution over all possible exit branches for a vehicle driving in a roundabout using recurrent neural network. Works such as \cite{Intentnet} and \cite{TNT}, use more complicated ML models to prediction vehicle's intention. Due to the complexity of the real-world and its ever-changing dynamics, the deployed autonomous vehicles inevitably face novel situations and should be able to cope with them. However, it has been repeatedly demonstrated that the reliability of ML models degrades radically when they are exposed to novel settings (i.e., under a shift away from the distribution of observations seen during their training) due to their failure to generalize, leading to catastrophic outcomes \cite{amodei2016concrete, sugiyama2012machine}. In \cite{OOD_AV}, Filos et al. also demonstrate that several state-of-the-art behavior prediction models for autonomous driving fail to generalize under arbitrary domain shift. Therefore, in this paper, we aim to address the domain generalization problem for vehicle intention prediction task and provide an effective solution to enhance reliability of autonomous vehicles. We hope our work can inspire further research in the field of improving generalizability for prediction algorithms in autonomous driving industry. 


\subsection{Methodology}
\subsubsection{Illustrate the problem context through a toy example}
\begin{figure}[htbp]
	\centering
	\includegraphics[scale=0.55]{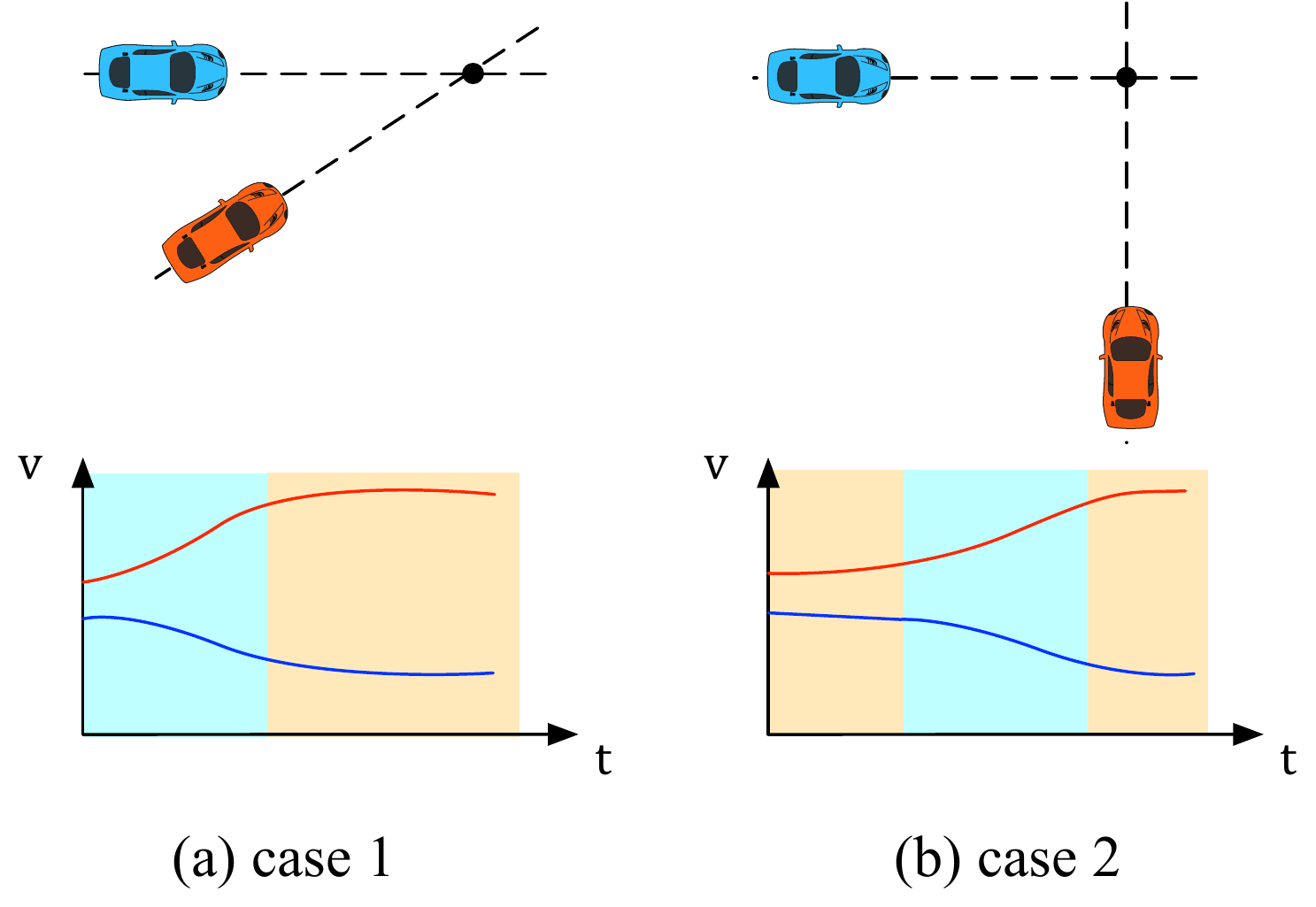}
	\caption{Illustration of our design motivation through a toy example.}
	\label{fig:motivation}
\end{figure}

Consider two vehicles interact with each other under two different domains. For the ease of understanding, we simplify the problem by temporarily making the following assumptions for the two interaction cases in Figure \ref{fig:motivation}: (1) The same road priorities; (2) The same two drivers interact with each other; (3) The same initial states for both vehicles. However,there are two major differences between two cases: (1) Different topological road structure of two intersected roads; (2) Two drivers have different driving styles (i.e. we let the driver in red car be more aggressive than the driver in blue car), which information is actually unavailable when making predictions. 

Based on previous assumptions and settings, for a given interaction period, a plausible simulated velocity profile of two vehicles is shown in the second row of Fig.~\ref{fig:motivation} for both cases. The goal is to predict the intention of any selected vehicle, which can be the intent of one car passes or yields the other car. According to the plots, whether the red car yields or passes the blue car doesn't rely much on the domain information but depends more on two drivers' personal information (e.g. driving style) and their initial states. Specifically, the geometry of the roads could affect several factors such as when two vehicles will start noticing each other, when the road negotiation will begin, and when drivers will agree on who should go first. All these factors could then influence the velocity profiles of two vehicles but won't have much effects on the intention of two vehicles when the aforementioned assumptions hold. In other words, if two vehicles encounter each other twice under different domains, as long as the three assumptions are true, the aggressive driver will always prefer passing than yielding the less aggressive driver. The intention is determined by features that relate to two drivers' information and their internal relations (e.g. highlighted in cyan), instead of those that relate to domain itself (e.g. highlighted in orange). Hence, we construct a model for the data collection process that assumes each input data (i.e. vehicle trajectories) is constructed from a mix of causal and non-causal features. We consider domain as a main intervention that changes the non-causal features of an input data, and propose that an ideal intention predictor should depends only on the causal features.

Although the aforementioned assumptions will not hold for real-world driving data, we are still able to assume each data is constructed from both causal and non-causal features. However, it becomes unclear what are the causal and non-causal features. Therefore, in this work, we construct a structural causal model and propose to learn an invariant representation for domain generalization based on invariant conditions identified from the causal model.

\subsubsection{Model Insights}
According to \cite{causal_review}, causality, with its focus on representing structural knowledge about the data generating process that allows interventions and changes, can contribute towards understanding and resolving some limitations of current machine learning methods. In fact, machine learning models that incorporate or learn structural knowledge of an environment have been shown to be more efficient and generalize better \cite{battaglia2016interaction, bapst2019structured, battaglia2018relational, santoro2017simple}. For human drivers, the way they interact with each other under different domains should also processes some invariant structural relations, since they can quickly adapt their driving skills to novel scenarios. Therefore, we construct a structural causal model (SCM) for vehicle intention prediction task to learn an invariant representation of input driving data for domain generalization. 

The domain generalization setting assumes the existence of domain-invariant patterns in the inputs (e.g. semantic features), which can be extracted to learn a predictor that performs well across seen and unseen domains. Moreover, causal predictive models are shown to have great generalizability since their output depends on stable relationships between input features and the outcome instead of associations between them. Therefore, in this work, we construct a structural causal model inspired from data generation process and propose to learn an invariant representation from temporal latent dependencies of input data for domain generalization, where the learning scheme is based on an invariant condition identified from the causal model. To be more specific, we assume each input data is constructed from a mix of inherent (causal) and domain-dependent (non-causal) features. Consequently, the captured temporal latent dependencies from sequential data also contain both causal and non-causal features. 

To generate a pair of interactive trajectories from two vehicles, data-generating process first samples two drivers' driving preferences ($O$) and a domain ($D$) that could be correlated to each other shown with dashed arrow in Fig. \ref{fig:framework} (e.g. driver behavior may vary from country to country ). Then, an interaction event ($E$) is sampled, which could correlate to domain information (e.g., the geometry of the roads may affect when two vehicles can notice/observe each other). The domain information is correspond to non-causal features ($X_{NC}$) and the driver information are correspond to causal features ($X_C$). Part of the event information, such as initial speed or location of two vehicles, are correspond to causal features which will influence the intention label. Another part of the event information, such as when the road negotiation begins, are correspond to non-causal features. Altogether, the causal and non-causal features construct the interactive trajectories ($X$) from where the time-related latent information ($Z$) are captured.

We further utilize D-separation to analyze the formulated structural causal model as it can: (1)connects causal relations and probabilistic independence; (2)computes all and only those independence and conditional independence relations that hold for all values of the parameter for any directed graph. At the lowest level of the causal hierarchy — association — we have discovered DAGs and d-separation as a powerful tool to reason about conditional (in)dependencies between variables.

\subsection{Related Theory}
\subsubsection{D-separation}
Here we provide some background concepts related to d-separation used in Section 2 for analyzing our causal graph. 

\textbf{Definition 1} \textit{\textbf{(d-connection)}: If G is a directed graph in which X, Y and Z are disjoint subsets of nodes, then X and Y are d-connected by Z in G if and only if there exists an undirected path U between X and Y such that for every collider C on U, either C or a descendent of C is in Z, and no non-collider on U is in Z. }

\textbf{Definition 2} \textit{\textbf{(collider)}: For any path between two nodes, a collider is a node where arrows of the path meet head-to-head.}

\textbf{Definition 3} \textit{\textbf{(d-separation)}: X and Y are d-separated by Z in G if and only if they are not d-connected by Z in G, denoted $(X \ind Y)|Z$.}

\begin{figure}[htbp]
	\centering
	\includegraphics[scale=0.45]{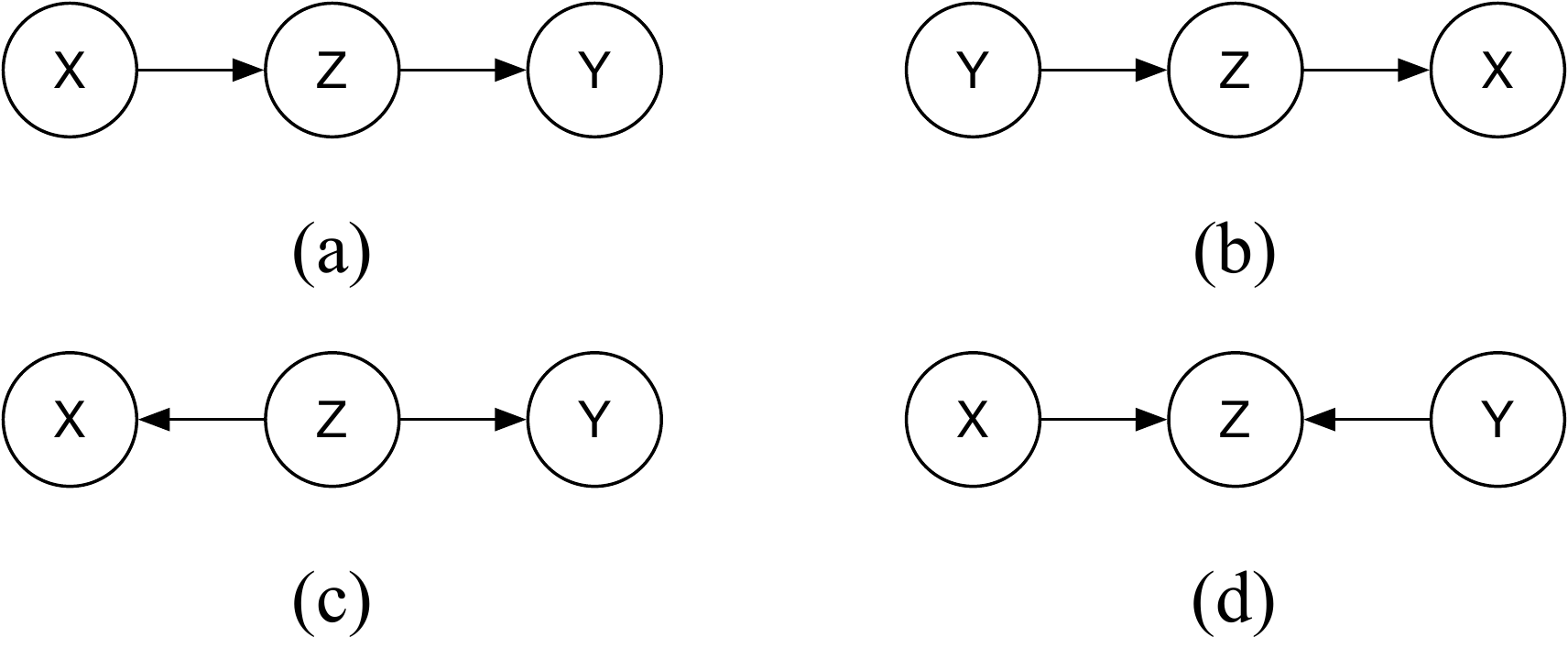}
	\caption{Four DAGs for three linked variables. The first two (i.e. (a) and (b)) are called chains; (c) is a fork; (d) is a collider. If these were the whole of the graph, we would have $X \not\ind Y$ and $X \ind Y|Z$. For the collider, however, we would have $X \ind Y$ while $X \not\ind Y|Z$.}
	\label{fig:causal}
\end{figure}

\subsubsection{Worse-case Generalization Error}
\normalfont
As mentioned in Suppl.\ref{related_works}, utilizing causal structures to
learn invariance factors has shown better model generalizability on unseen data and we also demonstrate desired performances of our causal-based model in Section 2. Theoretically (i.e. Theorem 1), the worse-case generalization error for a causal model is less than or equal to that for the associational model (i.e. model that directly map from X into Y). Detailed proof can be found in Suppl.A.2 in \cite{causal_2}.

\textbf{Theorem 1.} \textit{Consider a causal model $h_{c,D}^{\text{min}} : X_c \rightarrow Y$ and an associational model $h_{a,D}^{\text{min}} : X \rightarrow Y $ trained on a dataset $D \sim P(X, Y)$ with loss $L$. Let $(x, y) \in D$ and $(x', y') \notin S$ be two input instances such that they share the same true labelling function on the causal features, $y \in P(Y|X_c = x)$ and $y' \in P(Y|X_c = x')$. Then, the worst-case generalization error for the causal model on such $x'$ is less than or equal to that for the associational model. }
\begin{equation}
    \max_{x \in D, x'} L_{x'}(h_{c, D}^{\text{min}}, y) - L_{x}(h_{c, D}^{\text{min}}, y) \leq \max_{x \in D, x'} L_{x'}(h_{a, D}^{\text{min}}, y) - L_{x}(h_{a, D}^{\text{min}}, y)
\end{equation}

\subsubsection{Justification of Minimizing Contrastive Loss}

According to the invariance condition derived in Section 2.2, the idea function we want to optimize is $f^{*} = \operatorname*{arg\, min}_{h} \mathbb{E}[l(y, h(\mathbf{x}_c))]$ assuming $\mathbf{x}_c$ is known. However, as $X_C$ is unobservable, we need to alternatively minimize the objective function in Eq.(\ref{eq:L}). Inspired from Theorem 2 in \cite{MatchDG}, we are able to derive the following theorem under our problem setting. We can observe that whether $X_C$ will be returned by the optimization largely depends on $\delta_c$. Specifically, if a dataset has low $\delta_c$, then the learned representation will likely to be closer to the real $X_C$. Alternatively, if a dataset has high $\delta_c$, the matching condition loses any discriminative power and thus we need to additionally learn a matching function $\omega$ to minimize $\delta_c$. Therefore, minimizing the contractive loss (Eq.(\ref{eq:contrast_loss})) is necessary in our problem. Detailed proof or the original theorem can be found in Suppl.A.6 in \cite{MatchDG}.

\textbf{Theorem 2.} \textit{Assume training domains such that for any two same-class inputs $\mathbf{x}_j^{(k)}$ and $\mathbf{x}_j^{(k')}$ from domains $k$ and $k'$, $\text{dist}(x_{ac,j}^{(k)}, x_{ac,m}^{(k')}) \leq \delta_{ac}$ where $x_{ac}$ is any non-causal feature. Further, assume that the distance over $X_C$ between same-class inputs from different domains is bounded: $\text{dist}(x_{c,j}^{(k)}, x_{c,m}^{(k')}) \leq \delta_c$ and $\delta_c < \delta_{ac} (\delta_c, \delta_{ac} \in \mathbb{R}^{+})$. Then for some $\gamma$ and $\lambda$, a loss-minimizing classifier for the loss from Eq.(\ref{eq:L}) is the true function $f^*$, given a P-admissible loss function and a finite number of domains $k$ with $K \rightarrow \infty$ in each domain. }

\subsection{Experimental Details}
\normalfont

\subsubsection{Data Information}
All data are collected by a drone from bird’s-eye view with 10 Hz sampling frequency. Information such as reference paths and traffic regulations can be directly extracted from high definition maps that are in lanelet \cite{poggenhans2018lanelet2} format. The detailed information of each domain can be found in Table \ref{tab:domain_info}, where we observe that any two domains are different in at least two out of five aspects. The \textit{intersect angle} refers to the angle between two reference paths\footnote{A traffic-free reference path is normally obtained from road’s centerline.}. The \textit{average length} refers to the average length of two reference paths. 

Instead of Cartesian coordinate, we utilized the Fren\'{e}t Frame to represent vehicle's state. The vehicle motion in the Fren\'{e}t Frame can be represented with the longitudinal position along its reference path $s(t)$, and lateral deviation to the reference path $d(t)$. Therefore the vehicle state at time step $t$ can be defined as $\xi_t = (s(t), d(t))$. Note that the reference path of a vehicle will change according to the road it is current driving on. The origin of the reference path can be defined differently according to different objectives and each reference path will have its own Fren\'{e}t Frame. Since we are dealing with interaction between two vehicles, we define the origin as the intersection point (red circles in Fig. \ref{fig:nips_map}) of their reference paths.

\begin{table}[h!]
    \centering
    \caption{Detailed information of each domain.}
    \label{tab:domain_info}
    \begin{tabular}{c|c|c|c|c|c}
        \toprule
        & \multicolumn{2}{c|}{Road Topology} & Speed Limit & Traffic Rule & Location\\
        \midrule
        & intersect angle & average length & &  \\
        \midrule
        \midrule
        FT-1 &  $\approx 45^{\circ}$  & medium & 25 mph & yield sign & USA\\
        FT-2 &  $\approx 90^{\circ}$  & medium & 25 mph & stop sign & USA\\
        FT-3 &  $ > 90^{\circ}$       & short  & 25 mph & stop sign & USA\\
        ZS   &  $\approx 10^{\circ}$  & long   & 50 mph & zipper merging & China\\
        \bottomrule
    \end{tabular}
    
    \label{tab:my_label}
\end{table}

We manually select the interaction data from the dataset. Since we focus on two-vehicle interaction situations, we need to make sure the two vehicles are highly interactive with each other and both vehicles' behaviors should not be influenced by any other vehicles. Also, for ease of interpretation, we balance the data such that there are equal number of interactive trajectory data per class in each domain. We select a total number of 1354 time-series data from three domains in USA\_Roundabout\_FT and 138 time-series data from the single domain in CHN\_Merging\_ZS.

In terms of the labeling process, since we have the entire interaction trajectory of two vehicles, we are able to obtain the final outcome directly based on which vehicle passes the conflict point first. We categorize the intention for two interacting vehicles as either pass or yield and consider only the period that two vehicles have high interaction. In order to make sure that selected vehicle pairs are highly interactive and their behaviors are not influenced by other surrounding vehicles, we manually examine and filter the data. The are two reasons why we filter out the data that two vehicle are far from each other. Firstly, it is difficult to obtain true labels of drivers' intentions in such situation and their intentions may not simply be pass/yield. Secondly, if two vehicle are far from each other, there will be a large chance that they either haven't started interacting with each other or may interacting with other vehicles. It is worth to mention that our proposed approach can be adapted to multi-class classification with more subtle interaction patterns, but it might require human labeling.

\subsubsection{Implementation}
For domain generalization tasks, test/target domains are not accessible during training. Therefore, we randomly sample 20\% of data from source domains to construct our validation set which is used for hyperparameter tuning and use the remaining source domain data for training. We use the Adam optimizer and run all models for 500 epochs with learning rate of 1$e-$3. During training, we use a mini batch of 16 samples and apply early stopping criteria that the model does not experience a decrease in the validation loss for 20 epochs. The temperature scaling parameter $\tau$ is set to 0.05, and the two control parameters $\gamma$ and $\lambda$ are set to $1e-4$ and $1$, respectively.  
For results of each model, we report the average and standard deviation over 3 independent runs.

Input dimension of each data is $10 \times 4$, where we consider the past 1s of interactive trajectories with 10Hz sampling frequency and concatenate the states of two vehicles. Output classifer dimension is 2 corresponding to the possible number of intentions. Network modules in our model comprises: an encoder network $\varphi^{enc}_\tau$, which output parameters define the inference model $q(z_t^i|x_{< t}^i, z_{< t}^i)$ (this is also the representation function $q(\mathbf{x})$); a prior network $\varphi^{prior}_\tau$, which output parameters define the prior distribution $p(z_t^i|x_{< t}^i, z_{< t}^i)$; a decoder network $\varphi^{dec}_\tau$, which output parameters define the generative model $p(x_t^i|z_{\leq t}^i, x_{\leq t}^i)$; feature extraction networks $\varphi_\tau^x$ and $\varphi_\tau^z$ for $x$ and $z$, respectively; a recurrent process layer for hidden state update; a hypothesis function $\phi(z)$; and a function $h(\mathbf{x}_c)$ that maps causal features to the output intention. Specifically, all $\varphi_\tau$ consist of two hidden layers of 16 neurons with ReLU activation, where the input dimension is 4 and output dimension is 16. The hidden size of the recurrent layer is set to 16, where the input dimension is 32 and output dimension is 16. We consider $\phi$ to be the last layer of the network with one hidden layer of 16 neurons and ReLU activation, where the input dimension is 16 and output dimension is 2. We take $h$ to be identity function for simplicity. The dimension of latent representation $z$ is set to 2. When implementing state-of-the-art methods, since most of the baseline models are not designed for time-series input data, we add a RNN layer in front of each model instead of simply flattening the time-series input. The total number of parameters of all the baselines and our method are set to about equal.

\subsection{Additional Experiments and Results}

\subsubsection{Ablation Study}
We conduct an extensive study using both FT and ZS data to investigate contribution of three components to our method's performance: loss for extracting temporal latent dependencies, contrastive loss for learning matching function, and metrics to evaluate distance between two representations. To be more specific, {CTSDG w/o ${\mathcal{L}_v}$} is the CTSDG model without adding the loss term $\mathcal{L}_v$, where no temporal latent dependencies are modeled while learning invariant representations from the input data; {CTSDG w/o} ${\ell_{con}}$ is the CTSDG model without learning the matching function $\Omega$, where the loss $\mathcal{L}_r$ for satisfying invariance condition of $X_C$ is optimized by randomly selected input pairs; {CTSDG w/} ${s_{\ell_1}}$ and {CTSDG w/} ${s_{\ell_2}}$ are CTSDG models but using $\ell_1$ and $\ell_2$ as distance metric between two representations, respectively. From the ablation study results in Table \ref{tab:ablation_results}, we observe that by capturing the temporal latent dependencies and learning matching function for input pairs that share causal features, the model is able to have better performance especially when target domain has large domain shift compared to source domains (the fourth row). Moreover, according to the results from using $\ell_1$ and $\ell_2$ as distance metric, the cosine similarity used in our method seems to be the most suitable measure in the causal feature space for this prediction problem.

\begin{table}[h!]
  \begin{center}
    \caption{Ablation study of our method with prediction accuracy (\%).}
    \label{tab:ablation_results}
    \resizebox{\textwidth}{!}{
    \begin{tabular}{cc|cccccc}
      \toprule 
      Source & Target & CTSDG w/o $\mathcal{L}_v$ & CTSDG w/o $\ell_{con}$ & CTSDG w/ $s_{\ell_1}$ & CTSDG w/ $s_{\ell_2}$ & Ours \\
      \midrule 
      FT-1,2 & FT-3 & 75.79 (0.64) & 83.24 (3.87) & 81.75 (5.73) & 79.7 (3.18)  & 86.03 (1.48) \\
      FT-1,3 & FT-2 & 95.83 (0.39) & 95.15 (0.26) & 95.91 (1.28) & 96.08 (0.82) & 96.85 (0.15) \\
      FT-2,3 & FT-1 & 97.49 (0.63) & 97.45 (1.26) & 96.2 (1.90) & 95.19 (1.49) & 98.85 (0.13) \\
      FT-1,2,3 & ZS   & 77.29 (1.82) & 78.74 (3.02) & 80.67 (5.54) & 83.27 (4.74) & 85.51 (2.66) \\
      \midrule
      \multicolumn{2}{c|}{Average} & 86.60 & 88.65 & 88.69 & 88.56 & \textbf{91.81} \\
      \bottomrule 
    \end{tabular}}
  \end{center}
\end{table}



\subsubsection{Analysis on Invariant Representation}
\begin{figure}[htbp]
	\centering
	\includegraphics[scale=0.23]{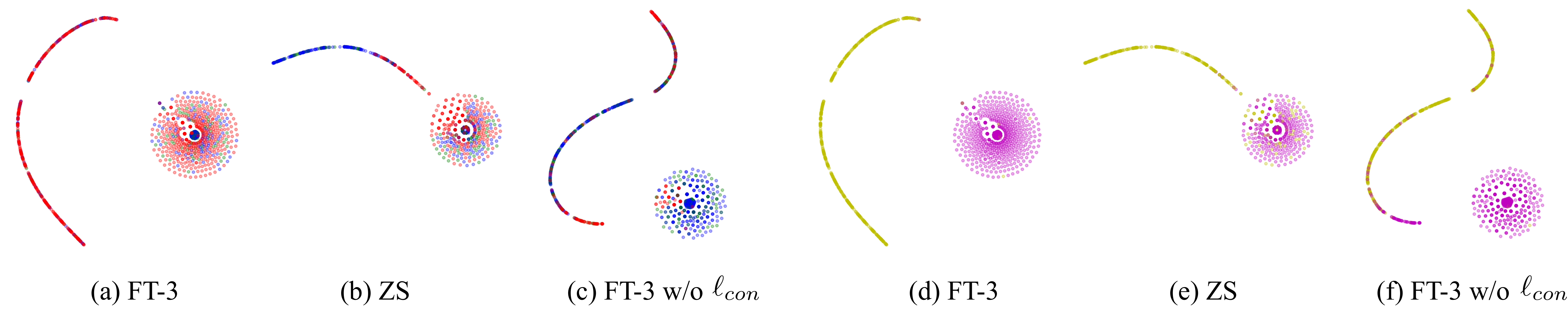}
	\caption{The t-SNE plots for visualizing latent representations. The source domains come from FT-1 and FT-2. The target domain for (a)(c)(d)(f) and (b)(e) is from FT-3 and ZS, respectively. For better visualization, we color (a)-(c) based on domain information: blue dots for FT-1, green dots for FT-2, and red dots for corresponding target domain. In (d)-(f), we color data points by ground-truth vehicle intention labels, where yellow dots represent \textit{pass} and magenta dots represent \textit{yield}.}
	\label{fig:tsne_results}
\end{figure}

We utilize t-SNE \cite{t-SNE} to analyze the quality of latent representation learned by our proposed model in Figure \ref{fig:tsne_results}. We train our model using two domains (FT-1,2) and evaluate on two unseen domains (FT-3 \& ZS). It appears from Fig. \ref{fig:tsne_results}(a) and (b) that CTSDG achieved a higher overlap between train and test domains for FT-3 than ZS, highlighting the difficulty of generalizing to the ZS domain. Such finding coincides with our prior domain knowledge as well as quantitative results in Table \ref{tab:FT_results}. Moreover, based on Fig. \ref{fig:tsne_results}(d) and (e), we find that the classes are well-separated in the training domains for both models. We also observe that when the intention label is \textit{yield}, the data is more evenly dispersed. The reason could because the input time-series data with \textit{yield} label are easier for the model to capture underlying domain representations that are invariant across time. Whereas the time-series data with \textit{pass} label could have more temporally diversed domain-invariant representations. 

In order to illustrate the effectiveness of learning match function $\Omega$, we further utilize t-SNE to compare the latent representation learned by CTSDG with that by CTSDG without the contrastive loss in Eq.~\ref{eq:contrast_loss}. Both models are trained using two domains (FT-1,2) and evaluated on an unseen domain FT-3. According to Fig. \ref{fig:tsne_results} (a) and (c), the model with learned match function has higher overlap between train and test domains than the model with random match function, especially when ground-truth target intentions are \textit{yield}. We can also observe from Fig. \ref{fig:tsne_results} (d) and (f) that without minimizing contrastive loss (Eq.~\ref{eq:contrast_loss}), target data from different classes cannot be well-separated in the latent representation, which generates miss-classification cases. Therefore, we can conclude that by enforcing the difference in individual representations for same-class inputs to be low through (Eq.~\ref{eq:contrast_loss}), the model can better capture domain-invariant representations.

\subsubsection{Analysis on Learned Temporal Latent Dependencies.}
\begin{figure}[htbp]
	\centering
	\includegraphics[scale=0.27]{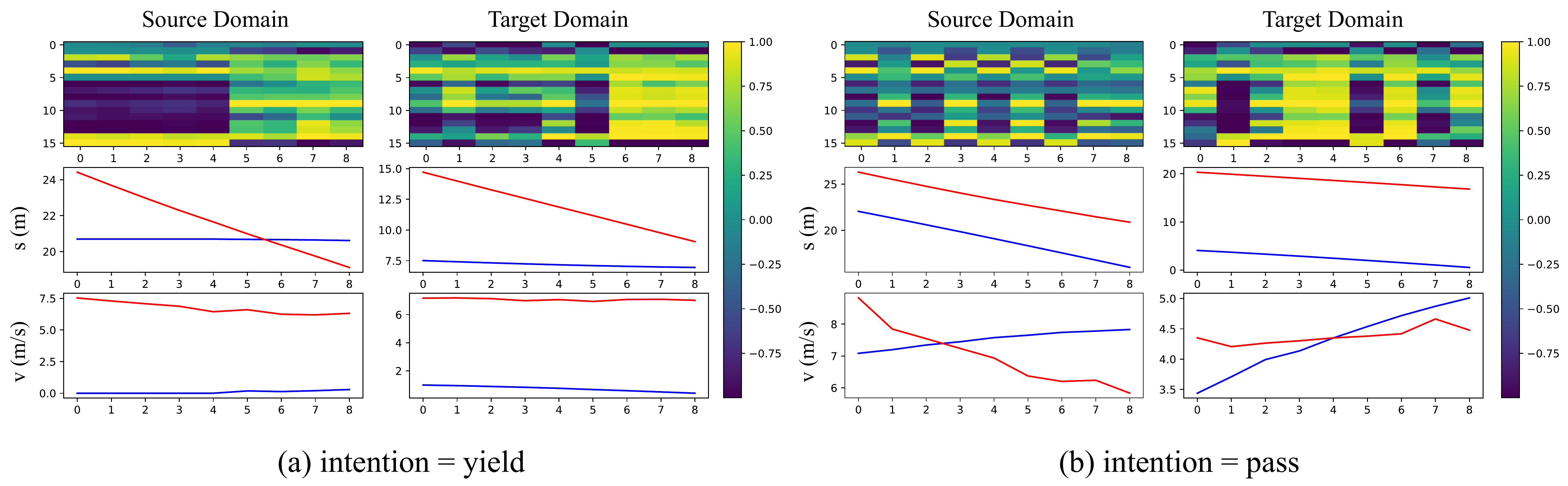}
	\caption{Learned temporal latent dependencies across time for different output intentions. The source domains are FT-1,2 and the target domain is FT-3. Each column shows a single sample from either source or target domain. The top row represents the cell state of memory cell for CTSDG, where y-axis refers to the activation of each neuron and x-axis refers to activation per time-step. The second row plots two vehicles' longitudinal coordinates under the Fren\'{e}t Frame with respect to time. The third row plots the velocity profiles for the two interacting vehicles.}
	\label{fig:vrnn_results}
\end{figure}

Figure \ref{fig:vrnn_results} shows the temporal latent dependencies learned by our CTSDG for a given source-target pair. We first look at the two cases from source domain (i.e. first column in Fig. \ref{fig:vrnn_results}(a,b)). Looking at the neuron activation patterns along the time axis, we can see that neurons are more active when there are changes in either position or velocity profiles for two interacting vehicles. In fact, these changes are related to causal features which play important roles in determining the output intentions based on our proposed causal model. As we look at the two cases from target domain (i.e. second column in Fig. \ref{fig:vrnn_results}(a,b)), we observe consistent excitatory neuron firing patterns at critical time steps, which indicate that CTSDG successfully transfers knowledge to unseen domains. Therefore, we can conclude that our CTSDG method is able to capture temporal latent dependencies that are domain invariant and related to causal features, which enables it generalize time-series driving data to unseen domains. 

\begin{figure}[htbp]
	\centering
	\includegraphics[scale=0.27]{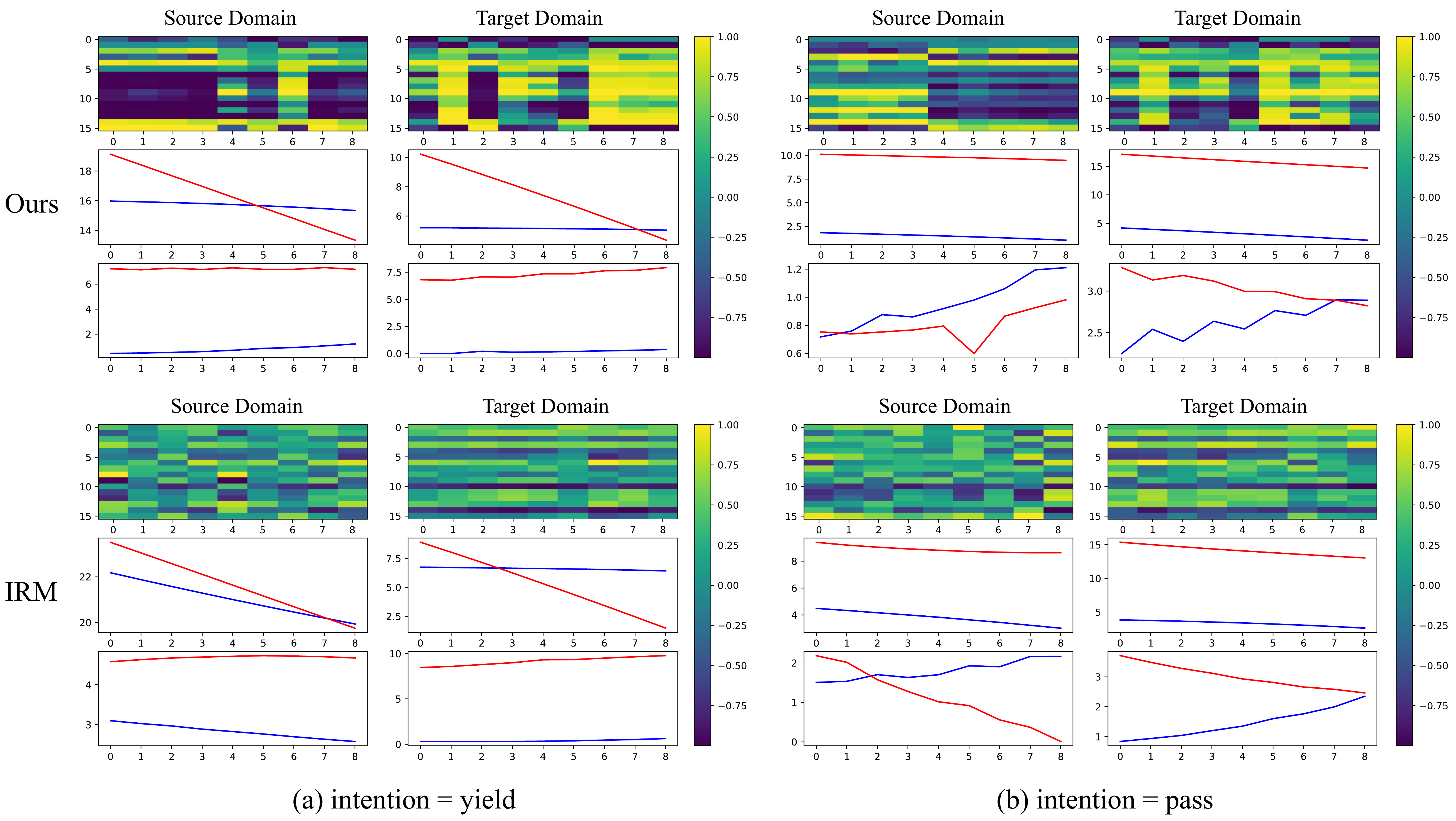}
	\caption{Learned temporal latent dependencies across time for different output intentions using different methods. The source domains are FT-1,2 and the target domain is FT-3. Each column under each method shows a single sample from either source or target domain. The top row under each method represents the cell state of memory cell, where y-axis refers to the activation of each neuron and x-axis refers to activation per time-step. The second row under each method plots two vehicles' longitudinal coordinates under the Fren\'{e}t Frame with respect to time. The third row under each method plots the velocity profiles for the two interacting vehicles. }
	\label{fig:vrnn_rnn}
\end{figure}

According to Table \ref{tab:FT_results}, IRM has the best performance among all baseline methods when the source domain is FT-1,2 and target domain is FT-3. Therefore we compare the temporal latent dependencies learned by IRM with those learned by our method (see Figure \ref{fig:vrnn_rnn}). We observe that IRM results in sporadic activation pattern and fails to capture and transfer temporal latent dependencies. In contrast, our method can extract critical temporal dependencies that are relevant towards two vehicles' intention and such dependencies can be further transferred from source to target domain.

\end{document}